\documentclass[runningheads]{llncs}
\usepackage{algorithm}
\usepackage[noEnd=true, indLines=true, italicComments=true]{algpseudocodex}
\usepackage{tikz}
\usepackage{standalone}
\usepackage{amsmath}
\usepackage{amsfonts}
\usepackage{xcolor}
\usepackage{subfig}
\usepackage{svg}
\usepackage{tabularx}
\usepackage{booktabs}

\newcolumntype{C}[1]{>{\hsize=#1\hsize\linewidth=\hsize\centering\arraybackslash}X}

\definecolor{nvidia}{rgb}{0.4627, 0.7255, 0}

\usetikzlibrary{arrows}
\usetikzlibrary{arrows.meta}
\usetikzlibrary{calc}
\usetikzlibrary{fadings}
\usetikzlibrary{fit}
\usetikzlibrary{matrix}
\usetikzlibrary{positioning}
\usetikzlibrary[positioning]

\usepackage[T1]{fontenc}
\usepackage{graphicx}
\begin{document}
%
%\title{Towards Efficient GPU-Accelerated Evaluation in GP-GOMEA}
\title{GP‑GOMEA with GPU‑Based Fitness Evaluations: Design and Performance Analysis}
\titlerunning{GP‑GOMEA with GPU‑Based Fitness Evaluations}
\author{Jasper Post\inst{1,2} \and
Johannes Koch\inst{1,2}\orcidID{0009-0008-7570-7621} \and
Anton Bouter\inst{1}\orcidID{0000-0003-4599-0733} \and
Tanja Alderliesten\inst{3}\orcidID{0000-0003-4261-7511} \and
Peter A.N. Bosman\inst{1,2}\orcidID{0000-0002-4186-6666}
}
\authorrunning{J. Post et al.}
\institute{Centrum Wiskunde \& Informatica, Amsterdam, The Netherlands \and
Delft University of Technology, Delft, The Netherlands \and
Leiden University Medical Center, Leiden, The Netherlands\\
}
\maketitle              % typeset the header of the contribution
\begin{abstract}
GP‑GOMEA is a state‑of‑the‑art evolutionary algorithm for symbolic regression, known for discovering small and interpretable models. However, its computational cost remains substantial, limiting its applicability to larger datasets and more complex target expressions. In contrast, the rise of modern subsymbolic approaches, particularly deep learning, has been driven largely by the massive parallelism offered by GPUs. In this work, we take the first major step toward a fully GPU‑accelerated GP‑GOMEA by introducing a GPU‑based fitness evaluation scheme. We design a GPU‑friendly representation of GP‑GOMEA’s template‑based individuals and a corresponding evaluation strategy that exploits the inherent parallelism of population‑based search. This substantially increases evaluation throughput, enabling orders of magnitude more evaluations within the same time budget. Across four standard symbolic regression benchmarks, this increased evaluation capacity yields performance improvements, particularly for larger datasets and larger population sizes. Moreover, the ability to efficiently evaluate much larger datasets and more complex templates enables analyses that were previously infeasible, allowing us to systematically analyze what makes expressions increasingly difficult for GP‑GOMEA, providing new insights into how expression structure affects search difficulty. Finally, for the first time, this expanded capability allows a problem‑agnostic evolutionary algorithm to reliably regress one of the largest Feynman equations within four hours.

\end{abstract}
\section{Introduction}
\label{sec:introduction}

% In many high-stakes domains such as healthcare and finance, predictive models must not only perform well but also be interpretable. Regulatory frameworks, including those proposed by the European Union, increasingly require that automated decision-making systems be transparent and explainable \cite{eu-xai}. Symbolic regression (SR) addresses this need directly by searching for explicit mathematical expressions that describe relationships in data, producing models that are inherently human-readable.

Symbolic regression (SR) aims to discover mathematical expressions that describe data. However, symbolic learning is known to be computationally demanding, especially for large datasets and complex expressions. Genetic programming (GP) is the dominant approach to SR, evolving candidate expressions through variation operators inspired by natural evolution.

Among GP‑based methods, GP‑GOMEA is one of the most effective symbolic regression algorithms, but its evaluation cost remains a major bottleneck. Meanwhile, modern machine learning routinely leverages GPU parallelism for large performance gains. This motivates bringing similar acceleration to GP‑GOMEA.

Since its introduction, several improvements to GP-GOMEA have been proposed. Despite these advancements, GP-GOMEA shares a bottleneck with most GP frameworks: fitness evaluation dominates runtime. This cost scales with both population size and dataset size, meaning that even moderately complex problems can require hours to solve \cite{srbench}, severely limiting the algorithm's ability to get good results on larger datasets.

In this work, we investigate whether the computationally intensive fitness evaluations in GP‑GOMEA can be offloaded to GPUs to reduce overall runtime. Parallelizing evaluation in GP‑GOMEA is non‑trivial because variation, evaluation, and selection are tightly interleaved. We therefore restructure the main evolutionary loop in GP-GOMEA to enable parallel fitness evaluations on the GPU. By exploiting modern GPU parallelism, we not only improve results for larger symbolic regression tasks but also gain the ability to perform analyses at expression sizes and complexities that were previously out of reach.

The remainder of this paper is organized as follows. Related work is covered in Section \ref{sec:related_work}. GP-GOMEA and the proposed changes are discussed in Section \ref{sec:method}. Experiments and results are covered in Section \ref{sec:experiments_and_results}. Finally, we discuss our results in Section \ref{sec:discussion} and conclude in Section \ref{sec:conclusion}.

\section{Related Work}
\label{sec:related_work}
The performance of symbolic regression algorithms is usually measured by the quality of solutions found within a fixed compute budget. Two complementary directions can improve scalability: algorithmic advances that reduce the number of evaluations needed, and computational optimizations that make each evaluation faster. We briefly review both, before situating our contribution.

GP‑GOMEA provides a substantial algorithmic improvement over standard tree-based GP by replacing random variation with linkage‑learning and gene‑pool optimal mixing, enabling much more effective search under strict depth limits~\cite{gp-gomea}. Since its introduction, several extensions have further improved its capabilities. One line of work addresses the handling of real-valued constants, a known weakness of GP-based methods, by proposing simultaneous evolution of constants and expression structure~\cite{gp-rv-gomea}. Another extends the algorithm to decomposable problems through the use of modular subexpressions~\cite{moduler-gp-gomea}. Further work improves efficiency for higher-arity operators by introducing a variation step that considers the semantic context of a subtree alongside greedy child selection, making better use of template nodes that would otherwise remain unused~\cite{higher-arity-gp-gomea}. More recently, improved linkage learning strategies based on node proximity have been shown to be both simpler and more effective than mutual information-based approaches~\cite{node_proximity}. While each advancement improves the search quality of GP-GOMEA, none address the computational cost of evaluation, which remains the dominant bottleneck. 

In symbolic regression, fitness evaluation normally requires evaluating a candidate expression on the full dataset. Consequently, the associated computational cost scales with both population size and dataset size. Reducing this cost is therefore a natural target for efficiency enhancement. A widely adopted approach is vectorized computation: frameworks such as Operon \cite{operon}, gplearn \cite{gplearn}, and others exploit SIMD instructions through libraries like Eigen or NumPy to evaluate expressions efficiently on a dataset. A complementary technique is Just-In-Time (JIT) compilation, used by PySR \cite{pysr} and Kozax \cite{kozax}, which specializes code at runtime, potentially yielding speedups at the cost of warm-up overhead.

On multi-core CPUs, the embarrassingly parallel nature of population-wide fitness evaluation in most forms of GP can be exploited through multi-threading or island-based models~\cite{pysr,deap}. GPUs provide massively higher parallel throughput, but their effective use requires different design choices. Existing GPU-based approaches vary in scope: SRGPU \cite{srgpu} offloads only evaluation and selection to the GPU, while EvoGP \cite{evogp} and Kozax \cite{kozax} execute all GP operations on the GPU. To support efficient GPU execution, tree representations are typically abandoned in favor of linear representations such as prefix notation (Polish notation) \cite{deap,srgpu,gplearn,evogp}, postfix notation (Reverse Polish notation) \cite{operon}, or reversed prefix \cite{kozax}, enabling stack-based evaluations without irregular control flow. 

Since GP‑GOMEA is already state‑of‑the‑art on a single core, bringing GPU acceleration to GP-GOMEA has potential for greater impact than when accelerating conventional GP approaches. So far, however, this has not been explored.

%We address this by restructuring the evaluation loop to enable batched parallel evaluation, and by introducing a GPU-friendly linear representation based on Reverse Polish notation. Crucially, these changes preserve the original algorithmic behavior of GP-GOMEA, including its linkage learning and optimal mixing mechanisms, meaning the GPU acceleration is fully compatible with existing and future extensions of the algorithm.
\section{Method}
\label{sec:method}
We first introduce the serial version of GP-GOMEA that we use as a basis for this work, followed by the GPU-accelerated version. 

\subsection{GP-GOMEA}
\label{subsec:gp-gomea}
As mentioned, several improved versions of GP-GOMEA exist to date. Here, we start from the recentmost published version~\cite{node_proximity}. While it does not include all improvements published so far, it incorporates what is arguably the most fundamental improvement so far, substantially enhancing the modeling and exploitation of linkage, which is key to any GOMEA variant.

GP-GOMEA iteratively refines a population of individuals (or solutions) through repeated variation and selection, similar to other EAs. Each individual in the population is represented as a fixed-length string of variables, mapped to nodes in a predefined tree template. This fixed structure ensures positional consistency across individuals, which is essential for identifying groups of statistically dependent nodes, referred to as linkage. Smaller expressions do not use all nodes in the template, leaving some variables inactive; these are called introns.
%While introns introduce redundancy, they are an accepted consequence of the fixed-template design.

Linkage is captured in a Family of Subsets (FOS), which contains multiple groups of linked nodes. While the FOS can be learned from population statistics, node proximity in the template has been shown to be a better source of linkage information~\cite{node_proximity}, and is used here. The Gene-pool Optimal Mixing (GOM) operator uses the FOS to guide variation: each individual is cloned to produce an offspring, and for each subset in the randomly shuffled FOS, the corresponding variables of the offspring are replaced with those from a randomly selected donor. The changed offspring is only evaluated if an actively used part of the template was affected. The change is accepted only if fitness does not worsen; otherwise it is reverted. If an individual fails to improve over $1 + \lfloor\log_{10}(\text{population size}) \rfloor$ subsequent generations, Forced Improvements (FI) are applied: GOM is repeated using the current best solution as the donor, stopping at the first improvement found \cite{parameterless-gomea}. If no improvement is found, the individual is replaced by a copy of the best solution. After GOM, the set of offspring replace the population.

Once the population converges, GP-GOMEA can be restarted: a new population is randomly generated, with the best solution found so far carried over to be used for FI. A restart is triggered after the No-Improvement Stretch (NIS), which is a number of subsequent generations in which the best solution did not improve. The default NIS is $\lfloor \log_{10}(\text{population size}) \rfloor$ generations.

% Don't use IMS in any experiment
% To avoid manual tuning of the population size, GP-GOMEA is typically combined with the Interleaved Multi-start Scheme (IMS) \cite{ims}. IMS maintains multiple populations of exponentially increasing sizes, evolved in an interleaved fashion: smaller populations perform generations more frequently, while larger populations are updated less often. Populations are terminated when they converge or are outperformed by a larger population, allowing the algorithm to automatically identify an appropriate population size.

\subsection{GPU-Accelerated Evaluations in GP-GOMEA}
Fitness evaluations are the computational bottleneck in single-core GP-GOMEA, especially for large population sizes and/or datasets. To accelerate GP-GOMEA, we offload evaluations to the GPU. Below, we describe how we achieved this.
%To do so, we restructure GOM, and adopt a memory layout and parallel decomposition optimized for GPU execution.

\subsubsection{Restructuring GOM.}
In the original serial GP-GOMEA~\cite{gp-gomea}, offspring undergo variation sequentially. For each individual, every subset in the FOS is considered in a random order and GOM is applied. Because of this, and because each variation operation can lead to a change in the best solution, which in turn may be used in FI, parallelization is not straightforwardly possible.

To enable parallel evaluation, we reversed the variation loop structure. In the parallel version, the outer loop iterates over the FOS, while the inner loop iterates over the population, with the FOS order shuffled per individual beforehand to preserve the stochasticity of the original implementation. Importantly, this only makes FI to be performed at a different point in time. In a recent version of GP-GOMEA that integrates real-valued evolutionary optimization, this was already found to not negatively impact the performance of GP-GOMEA~\cite{gp-rv-gomea}. For this reason, we used the same reversed-loop approach for both the CPU and the GPU version of GP-GOMEA in this paper.

Within the inner loop, in iteration $i$, variation is applied to each offspring, sequentially using the CPU, by copying the gene values pertaining to the $i$-th subset in the randomly shuffled FOS for that offspring, from a randomly selected donor. Instead of immediately evaluating the modified offspring, all modified offspring are collected into a batch. When all offspring have been changed using the CPU, the full batch of candidate offspring is evaluated in parallel on the GPU. Each candidate is then accepted if its fitness has not become worse; otherwise the modification is reverted. The resulting procedure is shown in Algorithm \ref{alg:gpu_gp_gomea}. The same loop reordering is also applied to FI, where the inner loop only considers the solutions to improve instead of the whole population.
\vspace*{-5mm}
\begin{algorithm}[ht]
    \caption{GPU GP-GOMEA\\
    \textit{Shown with FI condensed; parts executed in parallel on the GPU shown in \textcolor{blue}{blue}.}}
    \label{alg:gpu_gp_gomea}
    \small
    \begin{algorithmic}[1]
        \State $\mathcal{O} \gets \textsc{Clone}(\mathcal{P})$;  $\mathcal{B} \gets \textsc{Clone}(\mathcal{P})$
        \Comment{Offspring and Backup}
        \For{$i \in \{0, \dots, |\mathcal{O}|-1 \}$}
            \Comment{Shuffle FOS order for each offspring individual}
            \State $FOSOrder_i \gets \textsc{RandomPermutation}(|FOS|)$
        \EndFor
        \For{$i \in \{0,\dots,|FOS|-1\}$}
            \Comment{Loop over FOS indices}
            \State $changed \gets \{\}$
            \For{$j \in \{0, \dots, |\mathcal{O}|-1 \}$}
                \Comment{Loop over offspring}
                \State $donor \gets \textsc{Random}(\mathcal{P})$
%                \State $crossover\_mask \gets FOS[FOSOrder_{ji}]$
                \State $\mathcal{O}_j[FOS[FOSOrder_{ji}]] \gets donor[FOS[FOSOrder_{ji}]]$
                \If{$\mathcal{O}_j \neq \mathcal{B}_j$}
                    \State $changed \gets changed \cup j$
                \EndIf
            \EndFor
            \State \textcolor{blue}{$\textsc{Evaluate}(\mathcal{O}[changed])$}
            \Comment{Evaluate changed solutions on GPU (and update best)}
            \For{$j \in changed$}
%                \Comment{Acceptance}
                \If{$f(\mathcal{O}_j)$ not worse than $f(\mathcal{B}_j)$}
                    \State $\mathcal{B}_j \gets \mathcal{O}_j$
                    \Comment{Accept: update backup}
                \Else
                    \State $\mathcal{O}_j \gets \mathcal{B}_j$
                    \Comment{Reject: revert to backup}
                \EndIf
            \EndFor
        \EndFor
        \State $\textsc{PerformForcedImprovements()}$
        \State $\mathcal{P} \gets \mathcal{O}$
    \end{algorithmic}
\end{algorithm}
\vspace*{-10mm}

\subsubsection{Data Structure of Individuals}
The representation of individuals in GP-GOMEA (see Section \ref{subsec:gp-gomea}) is convenient for variation, but not for parallel evaluations on a GPU. In particular, evaluation requires recursive tree traversal, an operation GPUs handle poorly because it involves irregular control flow rather than uniform parallel work. To address this, all individuals are transformed into a GPU-friendly linear representation, as detailed below. The transformation cost is negligible compared to the benefits of GPU‑parallel evaluation.

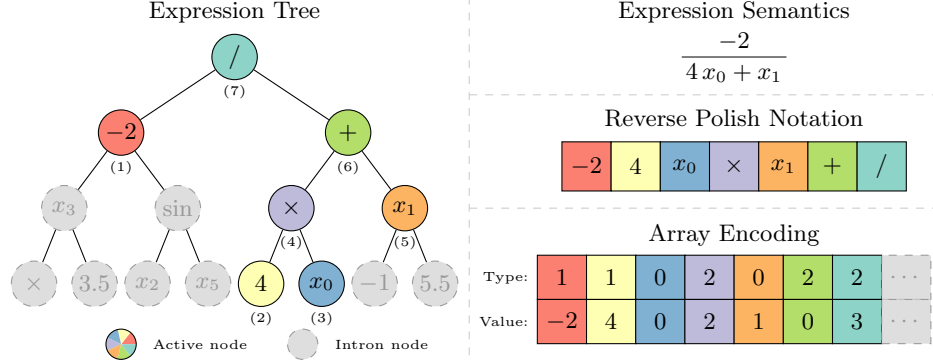
\begin{figure}[ht]
    \centering
    \definecolor{_light_blue}{RGB}{141,211,199}
\definecolor{_dark_blue}{RGB}{128,177,211}
\definecolor{_beige}{RGB}{255,255,179}
\definecolor{_purple}{RGB}{190,186,218}
\definecolor{_red}{RGB}{251,128,114}
\definecolor{_orange}{RGB}{253,180,98}
\definecolor{_green}{RGB}{179,222,105}
\definecolor{_intron}{RGB}{200,200,200}

\tikzset{
  active/.style={draw=black, circle, minimum size=6mm,
                 inner sep=0pt, font=\small},
  intron/.style={draw=black!40, dashed, circle, minimum size=6mm,
                 inner sep=0pt, font=\small, text=black!40},
  % suppress the 'every node' override inside child trees
  treenode/.style={},
}

\begin{tikzpicture}[
    level distance=10mm,
    level 1/.style={sibling distance=30mm},
    level 2/.style={sibling distance=15mm},
    level 3/.style={sibling distance=8mm},
]

  % ── Expression Tree ─────────────────────────────────────────────────────
  \node[active, fill=_light_blue] (n7) at (0,-0.2) {$/$}
    child { node[active, fill=_red] (n1) {$-2$}
      child { node[intron, fill=_intron!60] {$x_3$}
        child { node[intron, fill=_intron!60] {$\times$}   }
        child { node[intron, fill=_intron!60] {$3.5$} }
      }
      child { node[intron, fill=_intron!60] {$\sin$}
        child { node[intron, fill=_intron!60] {$x_2$} }
        child { node[intron, fill=_intron!60] {$x_5$} }
      }
    }
    child { node[active, fill=_green] (n6) {$+$}
      child { node[active, fill=_purple] (n4) {$\times$}
        child { node[active, fill=_beige]     (n2) {$4$}   }
        child { node[active, fill=_dark_blue] (n3) {$x_0$} }
      }
      child { node[active, fill=_orange] (n5) {$x_1$}
        child { node[intron, fill=_intron!60] {$-1$}  }
        child { node[intron, fill=_intron!60] {$5.5$} }
      }
    };

  % ── Post-order indices ──────────────────────────────────────────────────
  \foreach \nd/\idx in {n1/1, n2/2, n3/3, n4/4, n5/5, n6/6, n7/7}{
    % \node[draw=none, font=\tiny, below=1pt of \nd] {(\idx)};
    \node[draw=none, font=\tiny, anchor=north] at ([yshift=0.5mm] \nd.south) {(\idx)};
  }

  % ── Tree label ──────────────────────────────────────────────────────────
  \node[font=\small, draw=none] (tree_label) at (0, 0.4) {Expression Tree};

  % ── Legend ─────────────────────────────────────────────────────────────

  \begin{scope}[shift={(-1.5, -4)}]
    \foreach \col/\start/\end in {
      _red/0/51.4,
      _beige/51.4/102.8,
      _dark_blue/102.8/154.2,
      _purple/154.2/205.6,
      _orange/205.6/257.2,
      _green/257.2/308.4,
      _light_blue/308.4/360
    }{
      \filldraw[\col, draw=none]
        (0,0) -- (\start:2mm) arc (\start:\end:2mm) -- cycle;
    }
    \draw[black] (0,0) circle (2mm);
    \coordinate (la_east) at (2mm, 0);
  \end{scope}
  \node[font=\tiny, anchor=west] at ([xshift=3pt] la_east) {Active node};

  \node[intron, fill=_intron!60, minimum size=4mm]
    (li) at (0.9, -4) {};
  \node[font=\tiny, anchor=west] at ([xshift=3pt] li.east) {Intron node};

  % ── Dividing lines ───────────────────────────────────────────────────────
  % Placed at a fixed x between tree (spans ~ -3.5..3.5) and right column
  \draw[gray!40, dashed] (3.1, 0.6) -- (3.1, -4.2);
  \draw[gray!40, dashed] (3.1, -0.7) -- (9.23, -0.7);
  \draw[gray!40, dashed] (3.1, -2.2) -- (9.23, -2.2);

  % ── Right column ────────────────────────────────────────────────────────
  \def\rcx{6.6}  % x-centre of right column

  % Expression Semantics label
  \node[font=\small, draw=none] at (\rcx, 0.4) {Expression Semantics};

  % Equation
  \node[draw=none] (eq) at (\rcx, -0.2)
    {$\displaystyle \frac{-2}{\,4\,x_0+x_1\,}$};

  % RPN label
  \node[font=\small, draw=none] (rpn_label) at (\rcx, -1)
    {Reverse Polish Notation};

  % RPN matrix — use ampersand-replacement to avoid & conflicts
  
  \matrix (rpn) at (\rcx, -1.2) [
      matrix of nodes,
      nodes={
        draw=black, rectangle, minimum size=2em,        font=\small, row sep=-0.3em, anchor=center,
      },
      column sep=-\pgflinewidth,
      row sep=0pt,
      anchor=north,
  ] {
    |[fill=_red]|         $-2$  &
    |[fill=_beige]|       $4$   &
    |[fill=_dark_blue]|   $x_0$ &
    |[fill=_purple]|      $\times$   &
    |[fill=_orange]|      $x_1$ &
    |[fill=_green]|       $+$   &
    |[fill=_light_blue]|  $/$   \\
  };

  % ── Array Encoding ──────────────────────────────────────────────────────
  \node[font=\small, draw=none, anchor=north] (enc_label)
    at (\rcx, -2.3) {Array Encoding};

  \matrix (enc) at (\rcx, -2.7) [
      matrix of nodes,
      nodes={
        draw=black, rectangle,
        minimum width=2em, minimum height=1.8em,
        font=\small, inner sep=2pt, outer sep=0pt,
        anchor=center,
      },
      column sep=-\pgflinewidth,
      row sep=-\pgflinewidth,
      anchor=north,
  ] {
    % type row: 0=variable, 1=constant, 2=operator
    % -2: constant=1, 4: constant=1, x0: variable=0,
    %  *: operator=2, x1: variable=0, +: operator=2, /: operator=2
    |[fill=_red]|        $1$ &
    |[fill=_beige]|      $1$ &
    |[fill=_dark_blue]|  $0$ &
    |[fill=_purple]|     $2$ &
    |[fill=_orange]|     $0$ &
    |[fill=_green]|      $2$ &
    |[fill=_light_blue]| $2$ &
    |[draw=black!40, dashed, fill=_intron!60, text=black!40]| $\cdots$ \\
    % value row: constants→actual value, variables→index,
    %            operators: +=0, -=1, *=2, /=3, sin=4
    |[fill=_red]|        $-2$  &
    |[fill=_beige]|      $4$   &
    |[fill=_dark_blue]|  $0$   &
    |[fill=_purple]|     $2$   &
    |[fill=_orange]|     $1$   &
    |[fill=_green]|      $0$   &
    |[fill=_light_blue]| $3$   &
     |[draw=black!40, dashed, fill=_intron!60, text=black!40]| $\cdots$ \\
  };

  % Row labels, placed at the vertical centre of each matrix row
  \node[font=\tiny, draw=none, anchor=east]
    at (enc-1-1.west) {Type:};
  \node[font=\tiny, draw=none, anchor=east]
    at (enc-2-1.west) {Value:};

\end{tikzpicture}
    \vspace*{-0.5cm}
    \caption{Left: the expression tree; colored nodes are active, while gray dashed nodes are introns. Numbers below active nodes indicate post-order traversal indices. Top right: the expression encoded by the tree. Middle right: the same expression in RPN, with each node color-matched to its position in the tree. Bottom right: the array encoding used internally for GPU evaluation, storing node type (0\,=\,variable, 1\,=\,constant, 2\,=\,operator) and node value (variable index, constant value, or operator index). Dashed cells indicate zero-padding to arrive at a fixed array length.}
    \label{fig:eq_to_tree_to_rpn}
    \vspace*{-0.3cm}
\end{figure}

Individuals are transformed via post-order traversal to be represented in Reverse Polish Notation (RPN) \cite{operon}, where operators follow their operands (e.g., $3 + 1 \rightarrow 3\ 1\ +$) and introns are ignored. RPN encodes the evaluation order explicitly, allowing each subexpression to be evaluated before its parent operation. Evaluation can then proceed in a single forward pass using a stack. Importantly, this represents a sequence of simple operations that can be parallelized well on a GPU. Figure \ref{fig:eq_to_tree_to_rpn} illustrates an example. 

While RPN provides a representation that is easy and efficient to evaluate, the sequence must be encoded to make efficient use of memory. Following the approach proposed in \cite{evogp}, a single individual with an RPN sequence of length $L$ is represented using two arrays, $\mathbf{T} \in \mathbb{N}^L$ and $\mathbf{V} \in \mathbb{R}^L$, where $\mathbf{T}[i]$ encodes the discrete type of node $i$ (e.g., input, operator, constant), and $\mathbf{V}[i]$ stores the associated numerical value. A visual example is given in Figure~\ref{fig:eq_to_tree_to_rpn}.

Since the RPN representation of individuals can vary in length, each vector is padded to a fixed length $L_{\max}$, defined by the maximum number of nodes in template. This yields the modified arrays $\hat{\mathbf{T}} \in \mathbb{N}^{L_{max}}$ and $\hat{\mathbf{V}} \in \mathbb{R}^{L_{max}}$. Since all individuals now share the same length, they can be concatenated into two population-level arrays, $\mathbf{P}_{\hat{\mathbf{T}}} \in \mathbb{N}^{m \times L_{\max}}$ and $\mathbf{P}_{\hat{\mathbf{V}}} \in \mathbb{R}^{m \times L_{\max}}$, enabling efficient indexing and evaluation while preserving the benefits of RPN linearization.

% Division of work across threads and blocks
\subsubsection{Evaluation on the GPU}
In symbolic regression, fitness computation exhibits two levels of parallelism: across individuals in the batch (task parallelism) and across datapoints within an individual (data parallelism). The latter is only possible when each datapoint contributes independently to the loss, such as in the most commonly employed loss function: the Sum of Squares Error (SSE).
%Efficiently mapping sub-evaluations onto GPU hardware is critical for performance.  

We define a sub-evaluation as the computation of the squared error for one individual on one datapoint. The sub-evaluations are divided across threads, which are the parallel units of work executed on the GPU. Threads are organized into thread blocks, and executed in groups of 32, called warps, using single-instruction multiple-thread (SIMT) execution model, meaning that each thread in a warp executes the same instruction. If threads follow different control-paths, the execution is serialized, which reduces performance (this phenomenon is called warp divergence). To prevent this, each thread block is assigned to evaluate one individual. Since all threads in the block interpret the same individual, control flow remains identical across threads, avoiding warp divergence.
\begin{figure}[tbp]
    \centering
    \scalebox{0.96}{
    \input{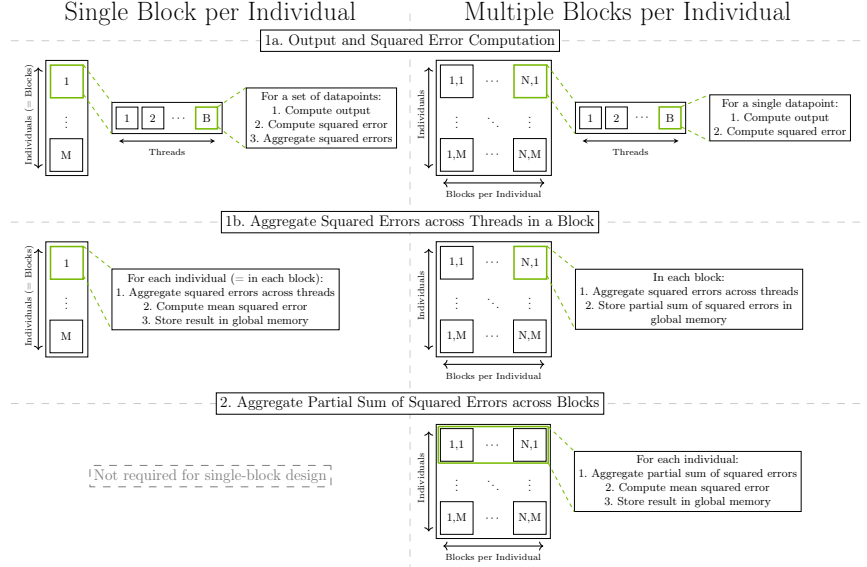}}
    \vspace*{-0.3cm}
    \caption{Parallel decomposition and kernel design for two designs. Left: single block per individual. Only requires a single kernel launch. Right: multiple blocks per individual. Requires a second kernel to aggregate partial sums of squared errors. M is the number of individuals to be evaluated, N is the number of blocks per individuals and B is the number of threads in a block.}
    \label{fig:pseudo_kernel}
    \vspace*{-0.25cm}
\end{figure}

The sequence data of an individual is stored in memory accessible only to threads in the same block, called shared memory, which is much faster than global memory. Although the data must be loaded from global memory at the start of the block's execution, this cost is offset by the faster access during subsequent accesses, increasing overall throughput.  

With up to 1024 threads per block, several parallelization strategies exist to cover all datapoints. (Figure \ref{fig:pseudo_kernel}): (1) let each thread process multiple datapoints sequentially, or (2) assign multiple blocks per individual and aggregate partial sums via global memory. The second option increases parallelism but adds global memory traffic and inter‑block synchronization. We therefore adopt the first strategy: each thread iteratively processes multiple datapoints and accumulates a partial sum in registers. Since all threads evaluating an individual reside in a single block, aggregation is performed within the same block.

Partial sums across threads are combined using parallel reduction, implemented via a CUDA provided primitive (\texttt{cub::BlockReduce}), which performs a tree-based reduction in shared memory. The resulting sum is normalized to compute the MSE and written to global memory for retrieval by the host.

\section{Experiments and Results}
\label{sec:experiments_and_results}

In this section, we evaluate the proposed GPU GP-GOMEA using three experiments. First, we benchmark GPU GP-GOMEA against its CPU counterpart. Secondly, we compare its evaluation throughput to several state-of-the-art symbolic regression approaches with both CPU and GPU implementations. Lastly, we demonstrate that GPU GP-GOMEA is capable of recovering an exact symbolic solution for a challenging benchmark problem that has proven difficult for prior GP-based methods, and investigate the sources of this difficulty.

\subsection{Experimental Setup}
The datasets used in the experiments are listed in Table \ref{tab:datasets}. The real-world datasets were selected as they are also used in \cite{evogp} and they cover a wide range of dataset sizes, allowing us to evaluate both solution quality and scalability. For Experiment I and II, each dataset was split into 30 cross-validation folds, where each fold uses $\frac{29}{30}$ of the instances for training; the remaining $\frac{1}{30}$ instances were used for validation. For Experiment III, each dataset was divided into 9 folds, where each fold uses $\frac{8}{9}$ of the instances for training, and the other instances were used for validation. In both experiments, each fold was assigned a different random seed. 

The Mean-Squared Error (MSE) is used as the objective:
\begin{equation}
\label{eq:mse}
    \text{MSE}(y_{\text{pred}}, y_{\text{target}}) = \frac{1}{N} \sum^{N}_{i=1}(y_{\text{pred},i} - y_{\text{target},i})^2
\end{equation}
where $y_{\text{pred}}$ denotes the predicted output, $y_{\text{target}}$ the target output, and $N$ the dataset size. However, to facilitate the comparison between datasets in Experiment III, the normalized mean-squared error (NMSE) is used:
\begin{equation}
\label{eq:nmse}
    \text{NMSE}(y_{\text{pred}}, y_{\text{target}}) = \frac{\text{MSE}(y_{\text{pred}}, y_{\text{target}})}{\text{Var}(y_{\text{target}})}
\end{equation}
where $\text{Var}(y_{target})$ is the variance of the target output.

\begin{table}[bt]
    \centering
    \caption{Datasets used in the experiments}
    \scalebox{0.93}{
    \begin{tabularx}{1.05\linewidth}{X c c c}
        \toprule
        \textbf{Dataset} & \textbf{Ground Truth Expression} & \textbf{\#Instances} & \textbf{\#Features} \\
        \midrule
        \multicolumn{4}{l}{\small\textit{Experiment I \& II datasets}} \\[2pt]
        Daily Demand \cite{daily_demand}      & $x_4 + x_5 + x_6$ & 60 & 12 \\
        Auto MPG \cite{auto_mpg}              & \textit{Unknown}   & 398 & 7 \\
        California Housing \cite{scikit-learn}& \textit{Unknown}   & 20{,}640 & 8 \\
        Feynman I.9.18 \cite{pmlb,feynman}   & $\frac{x_0\times x_1\times x_2}{(x_4-x_3)^2+(x_6-x_5)^2+(x_8-x_7)^2}$ & 100{,}000 & 9 \\[6pt]
        \multicolumn{4}{l}{\small\textit{Experiment III datasets}} \\[2pt]
        Addition       & $x_0+x_1+\cdots+x_8$ & 100{,}000 & 9 \\
        Division       & $\frac{x_0+x_1+x_2}{x_3+x_4+x_5+x_6+x_7+x_8}$ & 100{,}000 & 9 \\
        Subtraction    & $\frac{x_0+x_1+x_2}{x_4-x_3+x_6-x_5+x_8-x_7}$ & 100{,}000 & 9 \\
        Multiplication & $\frac{x_0\times x_1\times x_2}{x_4-x_3+x_6-x_5+x_8-x_7}$ & 100{,}000 & 9 \\
        Squaring       & $\frac{x_0\times x_1\times x_2}{(x_4-x_3)^2+(x_6-x_5)^2+(x_8-x_7)^2}$ & 100{,}000 & 9 \\
        \bottomrule
    \end{tabularx}
    }
    \label{tab:datasets}
    \vspace*{-0.5cm}
\end{table}

For comparison, we include CPU‑based GP-GOMEA \cite{node_proximity}, and GPU-based EvoGP \cite{evogp} and Kozax \cite{kozax}. We exclude Operon \cite{operon} from the main comparison because it requires linear scaling to be enabled when using MSE as the objective; since no other framework applies this transformation, including Operon would constitute an unfair advantage. Results with Operon are provided for reference in Appendix \ref{sec:framework_mse}. Our experimental setup follows that of \cite{evogp}, in which EvoGP was shown to perform competitively with several other symbolic regression frameworks, including TensorGP \cite{tensorgp}, PySR \cite{pysr}, DEAP \cite{deap}, SRGPU \cite{srgpu}, and gplearn \cite{gplearn}. Strong performance relative to EvoGP therefore provides an indication of competitive performance with respect to these methods, though we note that those comparisons were conducted under different experimental conditions.

To reflect typical usage of symbolic regression approaches, default parameter values are used whenever possible, except for the parameters explicitly listed in Table \ref{tab:experiment_setup}. Key hyperparameters, such as population size and template depth, are systematically varied, while the remaining parameters are kept at their default values (refer to Appendix \ref{sec:reproducibility}). The operator set was chosen to include commonly used mathematical operators for symbolic regression problems.

\begin{table}[tbp]
    \centering
    \caption{Experiment Setup}
    \label{tab:experiment_setup}
    \scalebox{0.93}{
    \begin{tabularx}{1.05\linewidth}{l|cc}
        \toprule 
         \textbf{Setting} & \textbf{Experiment I \& II (GPU \& CPU)} & \textbf{Experiment III (GPU only)}\\ 
         \midrule 
         Operator set & $\{+, -, \times, \div, \sin, \cos, {\cdot}^2, \sqrt{\cdot}, \exp, \log,\text{pow} \}$ & $\{+, -, \times, \div, {\cdot}^2, \sqrt{\cdot}\}$ \\
         Time budgets & \{10 min GPU, 60 min CPU\} & 240 min\\
         NISs & $\log_{10}$ & \{$\log_{2}$, $\log_{10}$\} \\
         Population sizes & \{128, 256, \ldots, 65,536\} & \{2,000, 65,536\} \\
         Template depths & \{4 (31 nodes), 6 (127 nodes)\} & \{6 (127 nodes), 8 (511 nodes)\}\\
         \bottomrule 
    \end{tabularx}
    }
    \vspace*{-0.5cm}
\end{table}

The difference between time budgets for CPU and GPU approaches in Experiments I and II reflects practical constraints in hardware availability, while remaining fair as GPU-accelerated methods can exploit SIMT parallelism to achieve substantially higher evaluation throughput. CPU-based approaches were executed on a single CPU core, while GPU-based approaches were executed on a single CPU core and a single GPU. We note that the CPU and GPU machines differ in their host CPU and available RAM (see Table \ref{tab:hardware}), which may modestly disadvantage GPU approaches beyond the time budget difference. For GP-GOMEA, populations are restarted when the population-wide NIS is reached, to ensure the full time budget is used. For the other approaches, default parameter settings are used, except that, following \cite{evogp}, the tournament size, crossover rate and mutation rate are set to 20, 0.9 and 0.1, respectively.
\begin{table}[tbp]
    \centering
    \caption{Hardware Specifications \textit{(note: CPU approaches use only a single core)}}
    \label{tab:hardware}
    \scalebox{0.93}{
    \begin{tabularx}{1.05\linewidth}{X|C{}C{}}
        \toprule
         \textbf{Component} & \textbf{CPU Approaches}  & \textbf{GPU Approaches} \\
         \midrule
         CPU & Intel Xeon E5-2699V4 & Intel Xeon Bronze 3206R  \\
         GPU & N/A & NVIDIA RTX A5000 \\
         System RAM & 630 GB & 96 GB \\
         \bottomrule
    \end{tabularx}
    }
    \vspace*{-0.5cm}
\end{table}

In Experiment III, restarts were performed using two separate configurations. The first configuration uses a starting population of 65,536 with a NIS of $\log_2(P)$ generations, favoring deeper exploration with a single population by allowing up to 16 generations without improvements before a restart. The second configuration uses a starting population of 2,000 with a NIS of $\log_{10}(P)$ generations, performing restarts more frequently. Running both configurations lets us assess whether GP‑GOMEA can exploit global landscape structure on harder problems, or whether many restarts perform better, which would be indicative of a highly multi-modal landscape with little global structure.

\subsection{Experiment I: CPU vs GPU GP-GOMEA}
\begin{figure}[bp]
    \centering
    \includegraphics[width=\linewidth]{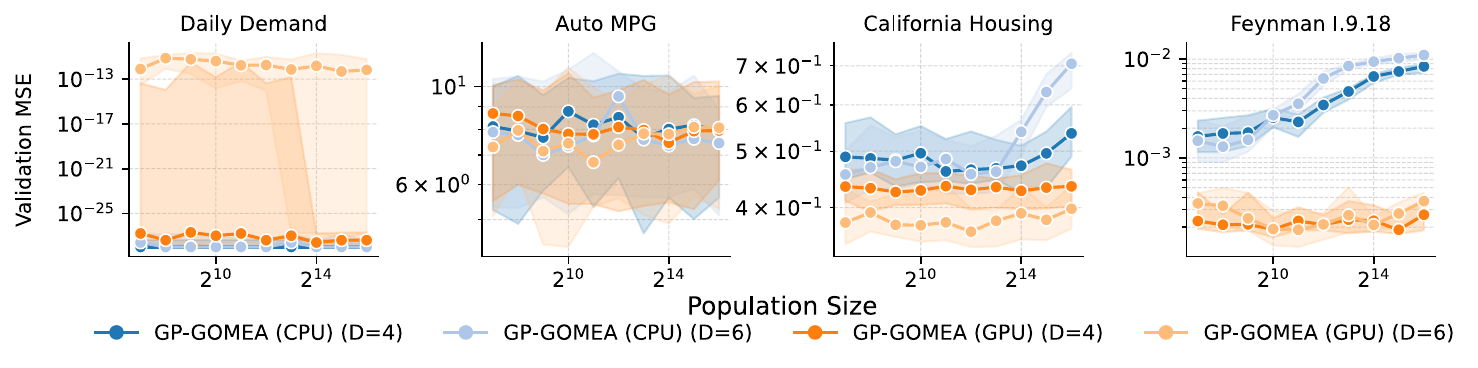}
    \vspace*{-0.75cm}
    \caption{Median validation MSE and interquartile ranges (shaded) over all runs for different population sizes across problems with template depths 4 and 6.}
    \label{fig:gomea_val_mse}
\end{figure}

\begin{figure}[htb]
    \centering
    \includegraphics[width=\linewidth]{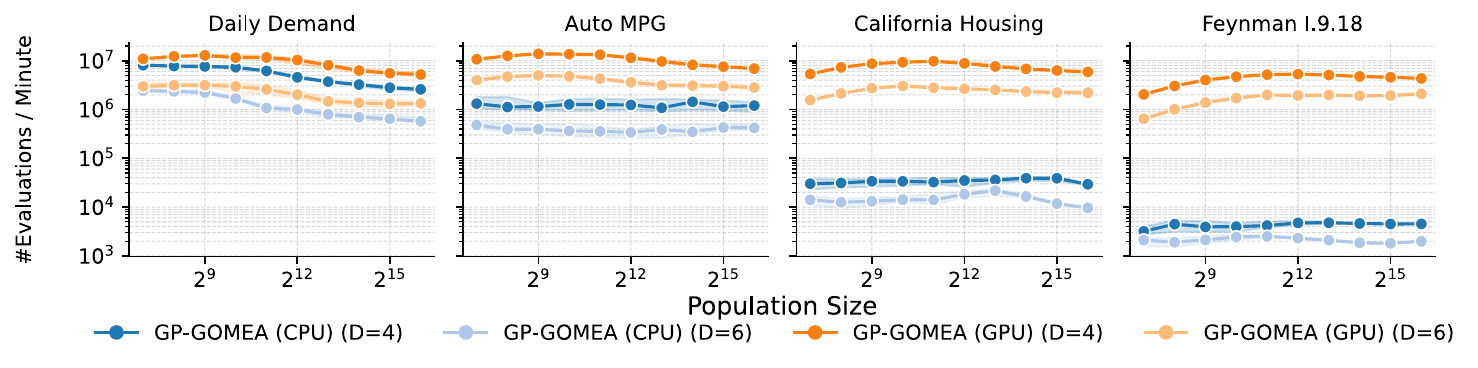}
    \vspace*{-0.75cm}
    \caption{Median number of evaluations per minute over population size across problems with template depths of 4 and 6. The shaded area corresponds to the interquartile range.}
    \label{fig:gomea_eval}
\end{figure}

We compare the best validation MSE obtained during a run (Figure~\ref{fig:gomea_val_mse}) and the number of evaluations per minute (Figure~\ref{fig:gomea_eval}). The MSE is recomputed from the returned symbolic expressions using SymPy \cite{sympy} with double-precision arithmetic, rather than relying on the values reported directly by GPU GP-GOMEA, ensuring fair and consistent comparisons. Training MSE results are provided in Appendix~\ref{sec:gomea_training_mse}.

\paragraph{Solution Quality (MSE).}
Across all datasets, GPU GP-GOMEA consistently matches or achieves a lower MSE than CPU GP-GOMEA, with the gap widening as dataset size increases. This trend is consistent with the evaluation throughput results discussed below.

\paragraph{Evaluation Throughput.}
The number of evaluations performed within the time budget provides insight into the effective search progress of each algorithm, since solution quality generally improves with more fitness evaluations. For implementations, evaluation throughput remains approximately constant as population size increases: larger populations reduce the number of generations, leaving the total evaluation count roughly unchanged.

GPU GP-GOMEA scales favorably with dataset size. On the smallest dataset (Daily Demand), its throughput is comparable to that of CPU GP-GOMEA. However, as dataset size increases, the benefits of GPU parallelism become increasingly pronounced. On the largest dataset (Feynman) GPU GP-GOMEA performs orders of magnitude more evaluations per minute than its CPU counterpart.

\subsection{Experiment II: Comparing Evaluation Throughput}
\begin{figure}[htbp]
    \centering
    \includegraphics[width=\textwidth]{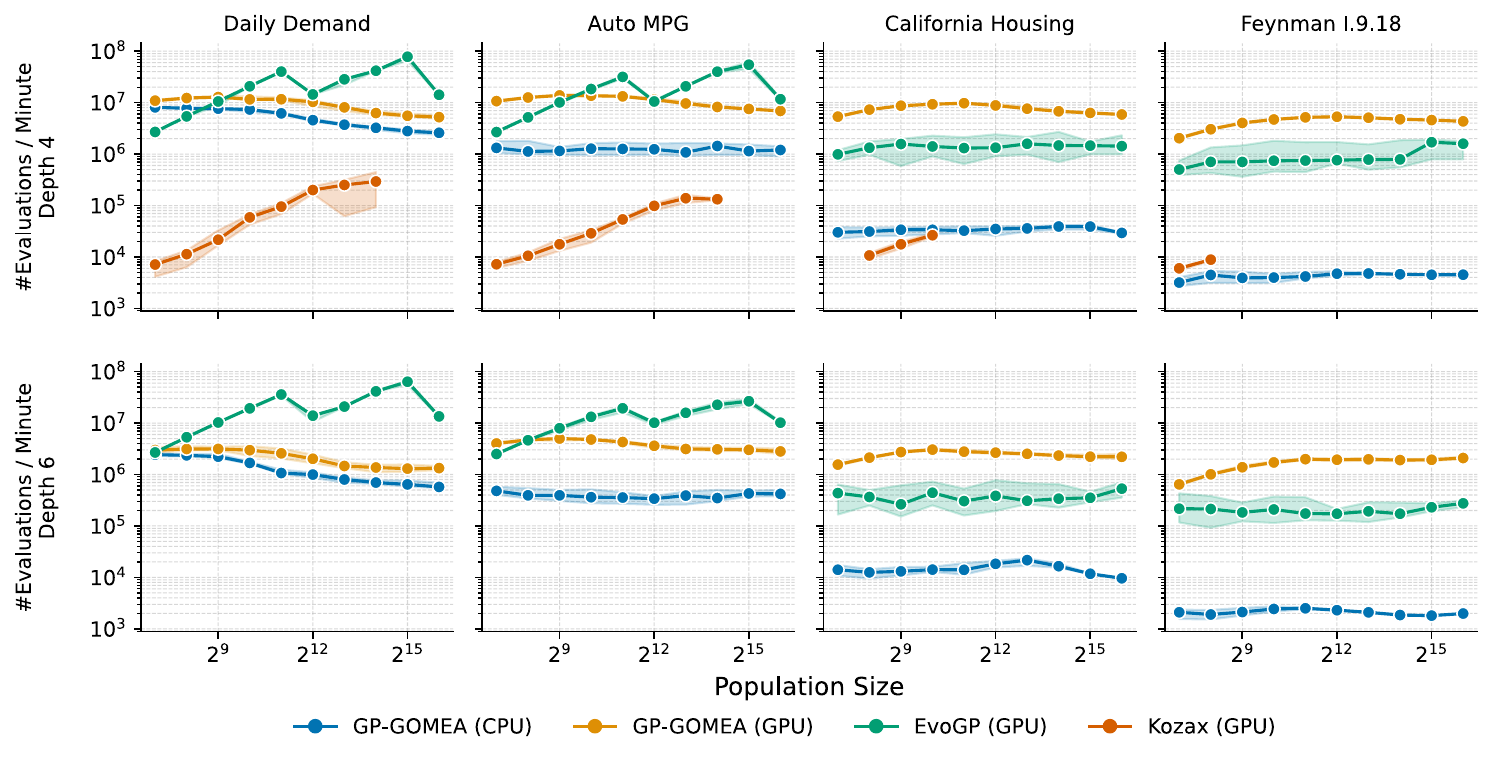}
    \vspace*{-0.75cm}
    \caption{Median number of evaluations per minute over population size across problems with template depths of 4 and 6. The shaded area corresponds to the interquartile range.}
    \label{fig:2_algo_eval}
    \vspace*{-0.5cm}
\end{figure}

We compare the number of evaluations per minute across frameworks (Figure~\ref{fig:2_algo_eval}). Since the frameworks have fundamentally different ways in which expression size is restricted, a direct accuracy comparison is complicated. An analysis of these differences, as well as MSE results, are provided for reference in Appendix~\ref{sec:framework_mse}. 

EvoGP achieves the highest evaluation throughput on the two smaller datasets (Daily Demand and AutoMPG), and its throughput increases markedly with population size on those problems, likely due to higher GPU utilization at larger population sizes. On small datasets, evaluation is extremely cheap and completes in microseconds, meaning the per-generation CPU overhead of GP-GOMEA's linkage model building and GOM becomes the dominant cost. EvoGP avoids this overhead entirely, as its simpler variation operators are implemented as GPU kernels, keeping the full evolutionary loop on the device. On the two larger datasets, however, GPU GP-GOMEA overtakes EvoGP in evaluation throughput, as EvoGP switches to launching a separate kernel per individual for large datasets, incurring per-launch overhead. 

Kozax performs significantly worse than all other methods in terms of evaluation throughput, executing several orders of magnitude fewer evaluations across most configurations, and additionally suffers from out-of-memory errors at template depth 6 and larger population sizes, limiting the configurations for which results are available.

\subsection{Experiment III: Finding a Large Feynman Equation}
The Feynman I.9.18 equation used in Experiment I represents a particularly challenging symbolic regression problem that has not been reported to be reliably solved using generic evolutionary symbolic regression methods. To the best of our knowledge, prior approaches reporting successful recovery rely on domain knowledge or recitation of memorized solutions in the case of neural and language models instead of pure search-based discovery~\cite{feynman,LLM-SR,QDSR}. Such approaches may not generalize to similarly difficult real-world problems.

\begin{figure}[tbp]
    \centering
    \includegraphics[width=\linewidth]{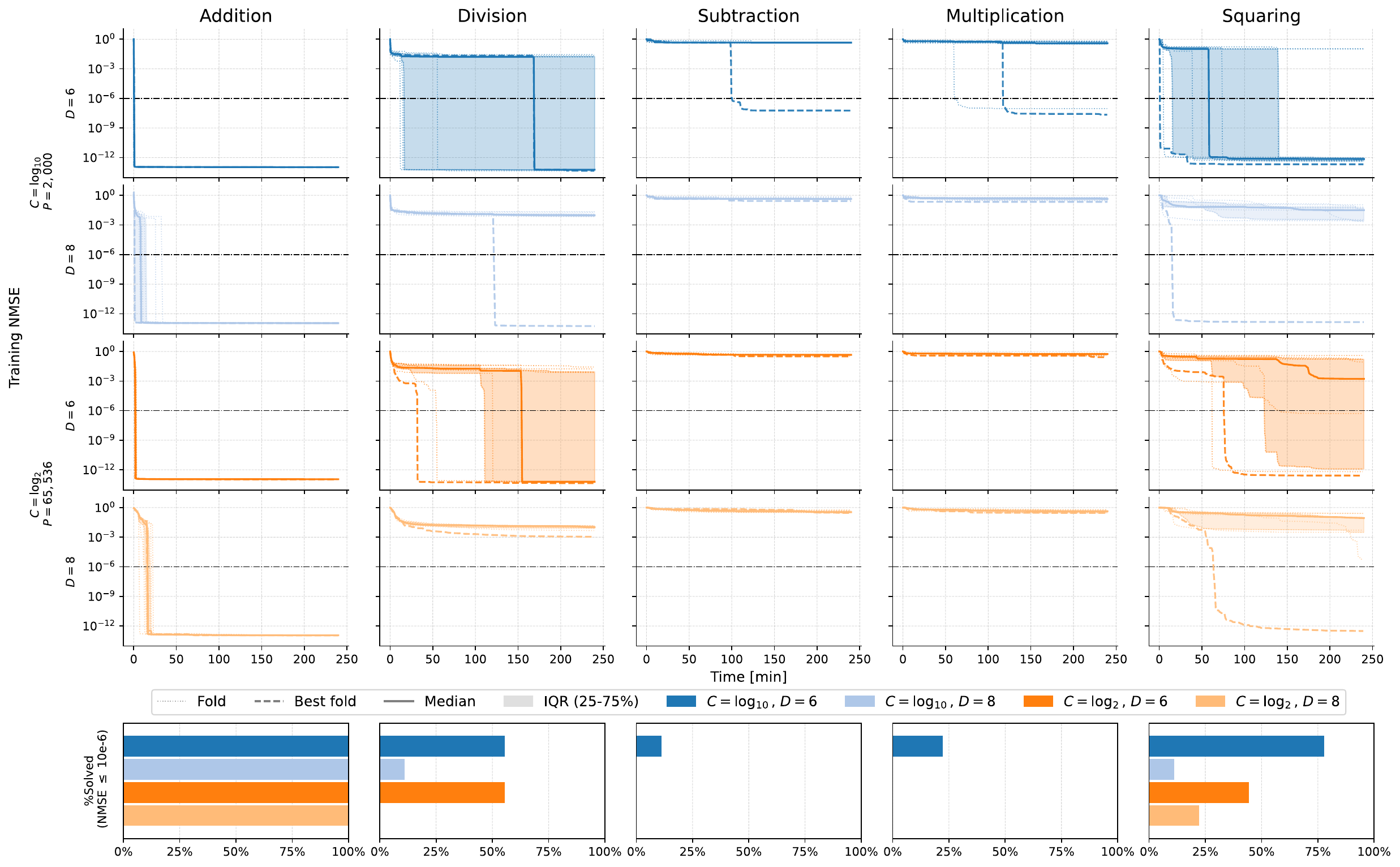}
    \vspace*{-0.5cm}
    \caption{Top rows: NMSE over time across problems with template depths of 6 and 8, and different restart configurations. Bottom row: success rate, defined as the percentage of folds that reached $\text{NMSE} \leq 10^{-6}$.} 
    \label{fig:3_gravity_nmse}
    \vspace*{-0.5cm}
\end{figure}

We want to see if GPU GP-GOMEA now offers enough efficiency to be able to find this equation. Beyond this, this efficiency enables us for the first time to also study the difficulty of this equation for symbolic regression (with GP-GOMEA) in more detail. Specifically, we evaluate GPU GP-GOMEA on a sequence of synthetic datasets corresponding to expressions of increasing complexity that progressively approach Feynman I.9.18 (see Table \ref{tab:datasets}). Performance is reported using the NMSE (Equation \ref{eq:nmse}) to facilitate comparison across datasets. Figure \ref{fig:3_gravity_nmse} shows the NMSE over time and the success rate, defined as the percentage of folds reaching $\text{NMSE} \leq 10^{-6}$. This corresponds to the effective decimal precision of 32-bit floating point arithmetic, and thus the tightest threshold at which NMSE loss remains a meaningful proxy for exact recovery rather than numerical noise.

\paragraph{Effect of Expression Complexity}
The Addition expression is recovered in all folds and configurations, though using a template depth of 8 requires more time due to the substantially larger search space.

As the expressions become more complex, a clear pattern emerges: the introduction of non-commutative operators is the primary driver of increased difficulty. While it is generally known and accepted that larger formulae are harder to find back, especially as they have more fractions and non-linear operators, our results confirm that non-commutative operators can make it harder as suspected in \cite{impact_commutative,maxwell1996why}. For the Division and Subtraction problems, discovery of the exact expression becomes increasingly unreliable. Both division and subtraction are non-commutative, meaning the order of operands matter, which reduces the number of global optima in the search space. Adding multiplication does not substantially change this, since multiplication is commutative and does not introduce additional ordering constraints.

Interestingly, the Squaring problem (i.e., the original Feynman equation), despite being structurally more complex than Subtraction and Multiplication, proves easier to solve. This is likely because squaring eliminates the ordering constraint on the subtraction operands, i.e.,: $(a - b)^2 = (b - a)^2$, making the subtraction effectively commutative within that context. This is supported by the success rate results in Figure \ref{fig:3_gravity_nmse}: success drops sharply when subtraction is introduced into the problem, and partially recovers when the square operator is introduced and renders the subtraction essentially non-commutative.

\paragraph{Effect of Template Depth and Convergence Settings}
A template depth of 6 leads to equal or better performance than a template depth of 8 across all problems. Although a deeper template can represent more complex expressions, it also dramatically increases the size of the search space, requiring substantially more time to converge. Since none of these problems require the representational capacity of depth 8, the additional complexity is counterproductive.

The difference between restart settings is less pronounced. However, the $\log_{10}(P)$ criterion with a starting population of 2,000 performs marginally better than the $\log_{2}(P)$ criterion with a starting population of 65,536. The more frequent restarts with smaller populations being beneficial could be indicative that these problems are highly multi-modal, where the local optima are likely largely disconnected with little exploitable global structure.

\section{Discussion}
\label{sec:discussion}
%This paper introduces GPU-accelerated evaluation in GP-GOMEA, substantially improving its efficiency while preserving its strong search capabilities.
%A central observation is that the performance gains are primarily driven by increased evaluation throughput.
The GPU implementation performs orders of magnitude more evaluations per minute, particularly on larger datasets. Since fitness evaluation dominates the computational cost in symbolic regression, this enables more extensive exploration of the search space and, consequently, improved solution quality. 

The experiments also show that the effectiveness of GPU acceleration is problem‑dependent, with the largest gains occurring on large datasets where enough parallelism potential exists to fully utilize the GPU.

Another notable observation is the impact of numerical precision. The GPU implementation relies on single-precision arithmetic, which can introduce small numerical errors compared to the double-precision CPU implementation. This becomes more apparent when expressions start to contain redundant subexpressions, as such subexpressions can amplify rounding errors, or redundant operations, which can even improve the NMSE.

For practical use, NMSE is not the only relevant criterion: users may reasonably prefer a less complex expression with slightly higher error. Moreover, highly complex expressions are more prone to overfitting, so focusing solely on the best‑NMSE solution is undesirable. For this reason, it is important for future work to bring the added value of GPU accelerations to a natively multi-objective version of GP-GOMEA. This would also help with avoiding redundant operations that improve the NMSE unexpectedly due to floating precision limitations.

In this paper, only evaluations are accelerated, while other components such as variation and linkage learning remain CPU-bound. Parallelizing these components could further improve scalability.

Finally, evolutionary real-valued optimization, modularity, and semantic inheritance were recently integrated into GP-GOMEA~\cite{gp-rv-gomea,moduler-gp-gomea,higher-arity-gp-gomea} and shown to lead to even better results. While it has not been tested whether GPU parallelization would have equally beneficial impact when all these facets, including multi-objective optimization, are integrated, we see no major limitations to do this and thereby obtain a highly practically efficient and effective GP-GOMEA version.
%extending the approach to additional features such as linear scaling \cite{linear_scaling} and multiple output trees \cite{multi-modal-gp-gomea} would require the parallel decomposition to be rethought and extensive modifications to be made to the current kernel design. Finally, assigning a single thread per individual can limit throughput when only a small number of individuals are evaluated, as the GPU is underutilized.
\section{Conclusion}
\label{sec:conclusion}
In this work, we investigated the use of GPU acceleration to address the computational bottleneck in GP-GOMEA: fitness evaluations. By restructuring the variation loop, introducing a GPU-friendly linear representation, and designing an efficient parallel evaluation strategy, we enabled effective utilization of GPU hardware without altering the core algorithmic principles of GP-GOMEA. 

Our results demonstrate that GPU GP-GOMEA significantly improves evaluation throughput, achieving up to a $600\times$ increase in evaluations per unit of time compared to the CPU implementation for larger datasets where GPU parallelism can be fully exploited. This enables GP‑GOMEA to scale its effective search to much larger expressions without bloating, due to the use of a template, and makes it possible, for the first time, to recover a very large expression (Feynman I.9.18) within a 4‑hour runtime in 78\% of runs.

% \input{mainmatter/example}
\bibliographystyle{splncs04}
\bibliography{report}

\newpage
\appendix

\section{Additional Experiment Results}
\label{sec:additional_results}

\subsection{Training MSE for GOMEA}
\label{sec:gomea_training_mse}

\begin{figure}
    \centering
    \includegraphics[width=\linewidth]{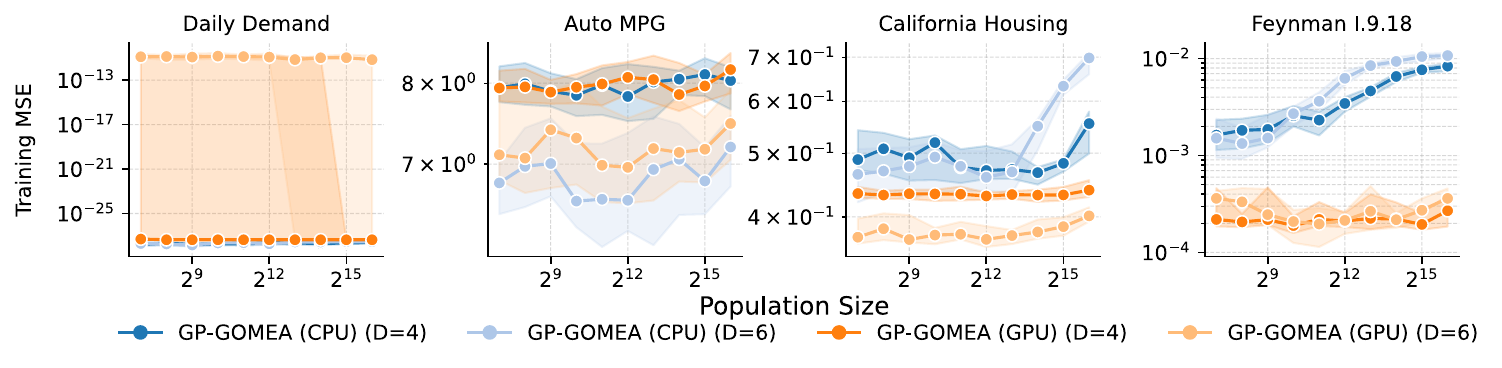}
    \caption{Median training MSE and interquartile ranges (shaded areas) over all runs for different population sizes across problems with template depths of 4 and 6.}
    \label{fig:gomea_train_mse}
\end{figure}

\subsection{Validation and Training MSE for GP Frameworks}
\label{sec:framework_mse}

\begin{figure}[htb]
    \centering
    \includegraphics[width=\linewidth]{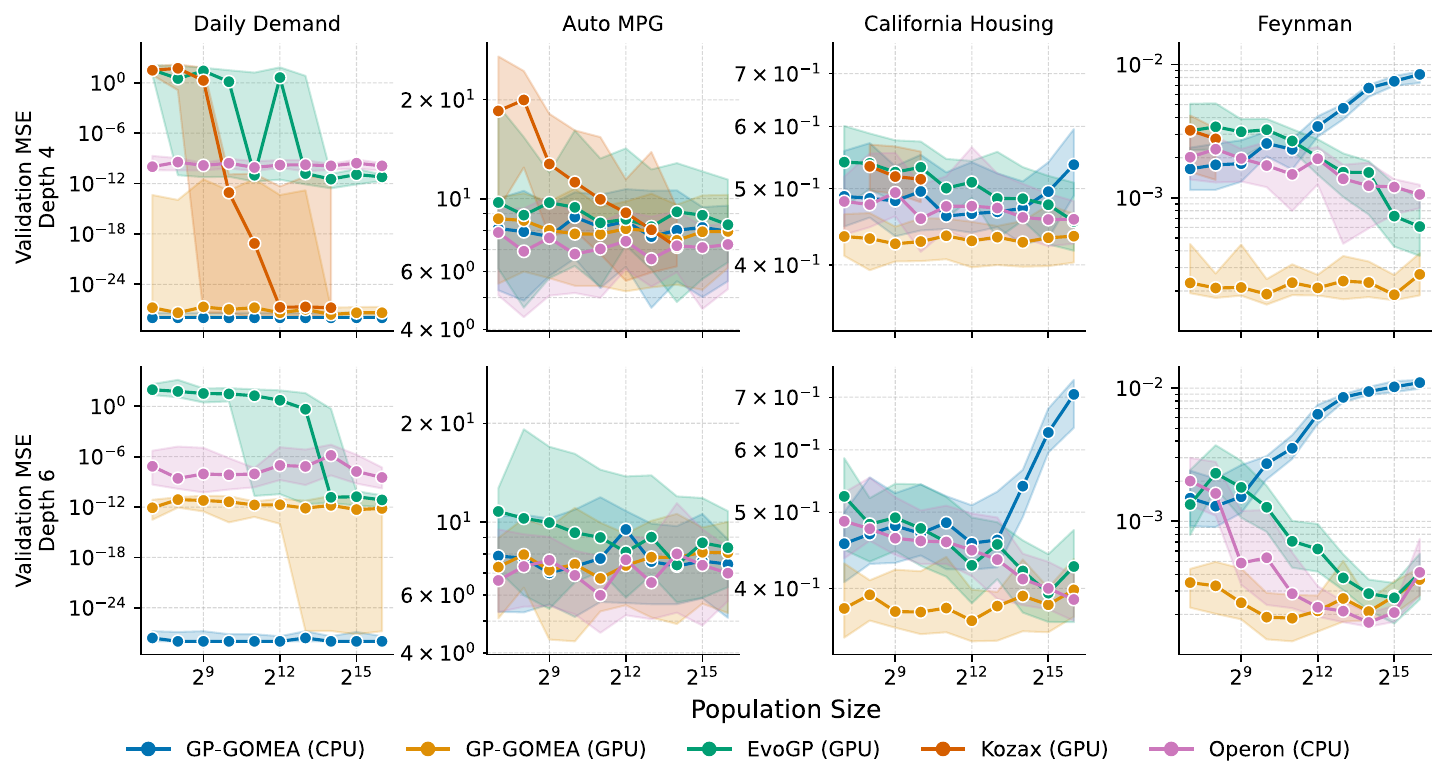}
    \caption{Median validation MSE and interquartile ranges (shaded areas) over all runs for different population sizes across problems with template depths of 4 and 6.}
    \label{fig:operon_val_mse}
\end{figure}

On Auto MPG, GPU GP-GOMEA is narrowly outperformed by Operon (see Figure \ref{fig:operon_val_mse}), likely because Operon employs linear scaling by default and has many coefficients hard-coded in the expression (one per tree node). While linear scaling has been implemented for GP-GOMEA \cite{node_proximity}, it is not yet supported in combination with the GPU evaluation introduced in this work. This limitation is mostly relevant for datasets where the underlying expression is unknown and may include coefficients (Auto MPG and California Housing).

\begin{figure}[htb]
    \centering
    \includegraphics[width=\linewidth]{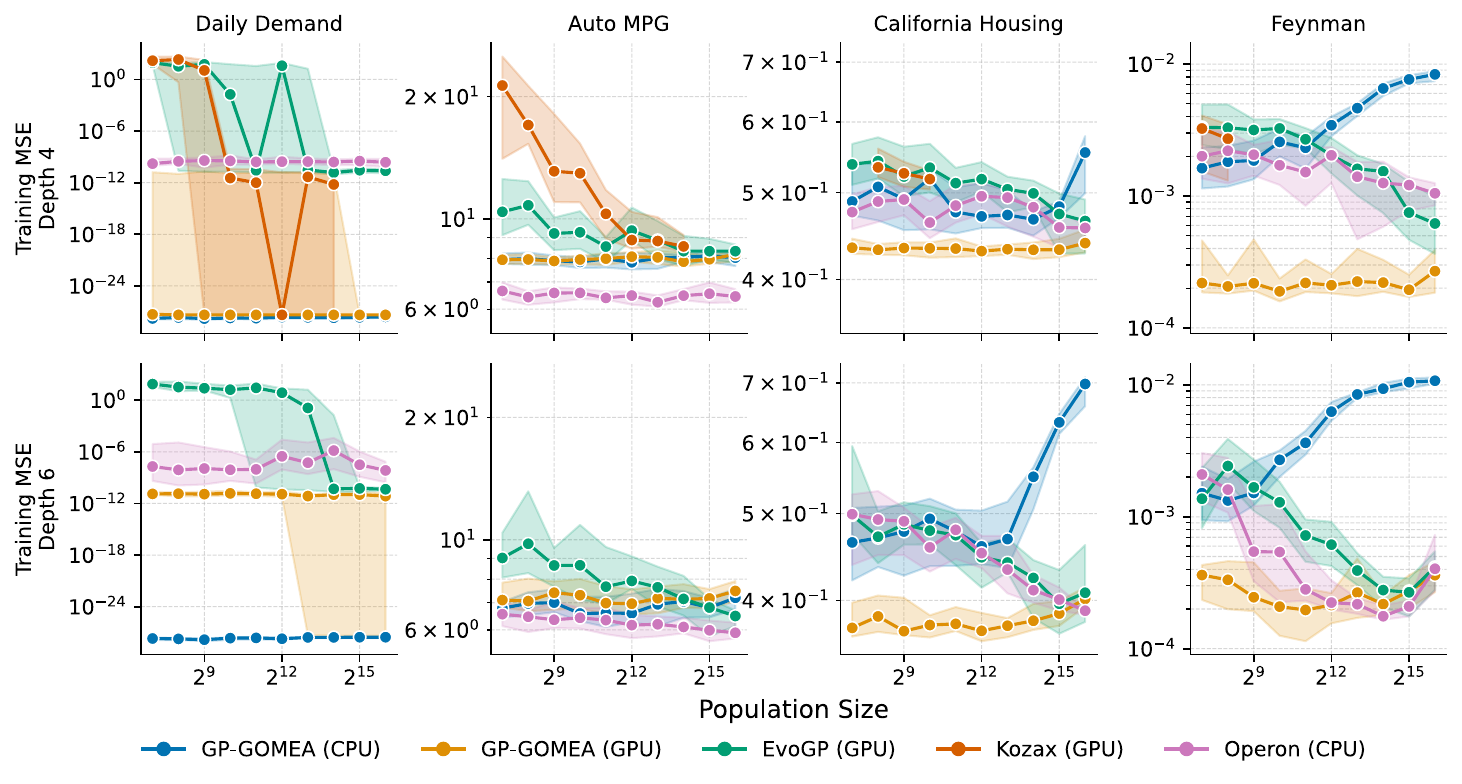}
    \caption{Median training MSE and interquartile ranges (shaded areas) over all runs for different population sizes across problems with template depths of 4 and 6.}
    \label{fig:operon_train_mse}
\end{figure}

\paragraph{Algorithmic Differences of Frameworks}
A direct comparison of validation and training MSE across frameworks is complicated due to how each framework enforces structural constraints on expression trees. GP-GOMEA guarantees that a tree never exceeds the specified depth limit during variation. The other frameworks, however, do not uphold this guarantee. While trees are initialized with either a maximum depth, maximum number of nodes, or both, the depth constraint is not enforced during variation, only the node limit is. As a result, the expression trees produced by these frameworks can represent a substantially wider class of expressions than those produced by GP-GOMEA, constituting a methodological inconsistency that makes direct MSE comparisons unreliable.

This is illustrated by the expressions with the lowest MSE found for the Auto MPG dataset at a specified depth of 4 (maximum of 31 nodes) by EvoGP, Kozax, and Operon respectively (coefficients are rounded to 2 decimal places):

\begin{equation}
    3.88\sqrt{\dfrac{x_5\left(x_5 + 39.64\cos^4(6.52\, x_5)\right)}{x_0 - (x_0 - 9.07)\cos^2\!\left(x_3^{1/4}\right) + 82.31}}
\end{equation}
\begin{multline}
    \log^2\!\left(\frac{x_5}{x_4}\right) + \sin\!\left(\log^2\!\left(-x_0 + 2\left(-x_5 + \sin(x_5)\right)^2\right)\right) +\\ \frac{x_4\left(x_4 - x_5 + \sin\!\left(x_5^2\right)\right)^2}{x_3}
\end{multline}
\begin{multline}
    2.90 + 8.37 \cdot \left(e^{\,e^{\dfrac{\log(1.29)}{e^{\sin(-0.49\, x_5)}}}} + \cos\!\left((0.24\, x_5)^2 - 0.63\, x_5\right) + 0.24\, x_5\right) \times \\ \dfrac{0.36\, x_6 + 0.24\, x_5 + \cos\!\left((0.46\, x_2)^2 - \sin(6.26\, x_4)\right)}{0.45\, x_2 + 0.05\, x_3}
\end{multline}

All three expressions require tree depths well in excess of the specified limit of 4. Additionally, Operon applies linear scaling and assigns an individual coefficient to every variable, meaning it violates not only the depth constraint but also the node limit. EvoGP and Kozax, by contrast, do adhere to the node limit.

For reference, the expressions found by the CPU and GPU implementations of GP-GOMEA for the same dataset and depth limit are:  

\begin{equation}
    \frac{x_5 - 25.15}{\sqrt{19.58}} + \left(\frac{-33.86}{31.63}\right)^{x_1 x_6} + \frac{x_5^2 - (x_3 + x_2)}{(x_2 + 44.79) + (x_0 + x_6)}
\end{equation}

\begin{equation}
    \frac{\dfrac{x_5^2}{\sqrt{x_3}}}{\cos(-26.91) + \log(x_2)} + \cos(-6.83\, x_5) - \frac{\log(x_4)}{x_6}
\end{equation}

These expressions are visibly more compact and conform strictly to the imposed depth limit, reflecting GP-GOMEA's stricter enforcement of structural constraints throughout the search process.

\subsection{Evaluation Throughput for GP Frameworks}
\label{sec:eval_gp_}
\begin{figure}[htb]
    \centering
    \includegraphics[width=\linewidth]{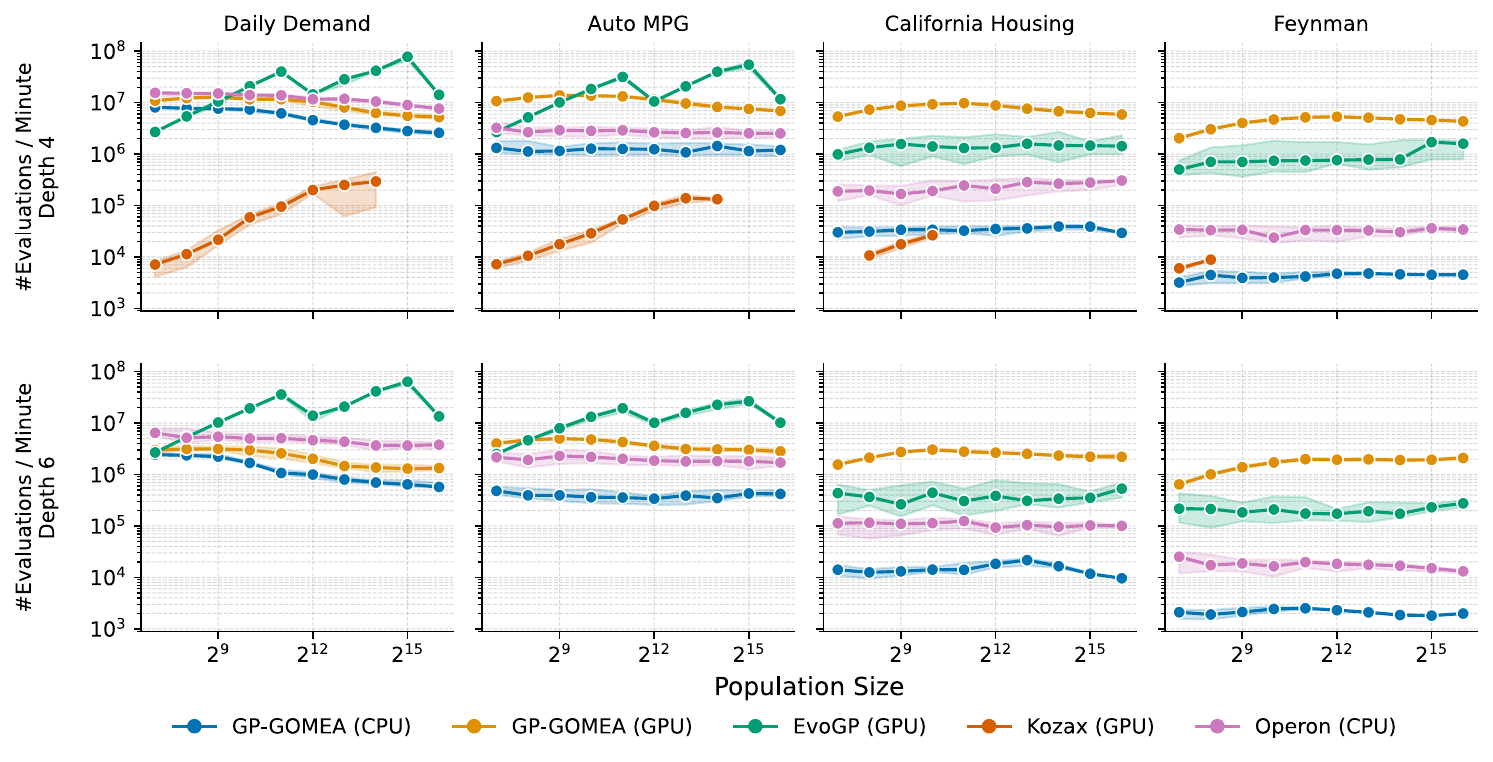}
    \caption{Median number of evaluations per minute and interquartile ranges (shaded areas) over all runs for different population size across problems with template depths of 4 and 6.}
    \label{fig:operon_evaluation}
\end{figure}

Operon consistently outperforms CPU GP-GOMEA in throughput and scales similarly, but is only competitive with GPU GP-GOMEA on the smallest dataset (see Figure \ref{fig:operon_evaluation}). 

\newpage
\section{Reproducibility and Software Environment}
\label{sec:reproducibility}

To facilitate reproducibility, this section provides details on the software environment and hyperparameter configuration used in all experiments.

\paragraph{Software Versions}
Table \ref{tab:software_versions} lists the specific versions and commit hashes of all the evaluated frameworks. EvoGP did not include a dedicated evaluation counter, so the project was forked to add this functionality\footnote{https://github.com/JRPost11/evogp}.

\begin{table}[htb]
    \centering
    \caption{Software versions and commit hashes for all evaluated frameworks.}
    \label{tab:software_versions}
    \begin{tabular}{l|c}
        \toprule 
        \textbf{Framework} & \textbf{Version/Commit Hash} \\
        \midrule
        EvoGP & 19b597130df6b53e3b6cbc0e66ad2d30374f1b69 \\
        Kozax & 0.0.11 \\
        Operon & 0.6.0 \\
        \bottomrule %\hline
    \end{tabular}
\end{table}

\paragraph{Hyperparameters}
Hyperparameters not explicitly discussed in Section \ref{sec:experiments_and_results} were kept at their default values. Since default values may change across future versions, the version number and commit hashes in Table \ref{tab:software_versions} serve as a stable reference for reproducing the exact configuration used here.





% LNCS Homepage, \url{http://www.springer.com/lncs}, last accessed 2023/10/25
% \end{thebibliography}
\end{document}